\documentclass[journal,transmag]{IEEEtran}

\usepackage{graphicx}
\usepackage{apalike}
\usepackage{multirow}
\usepackage[table]{xcolor}
\usepackage{algpseudocode}
\usepackage{algorithm}
\usepackage{amsmath}
\usepackage{dsfont}

\begin{document}
\algnewcommand\algorithmicforeach{\textbf{for each}}
\algdef{S}[FOR]{ForEach}[1]{\algorithmicforeach\ #1\ \algorithmicdo}

\hyphenation{op-tical net-works semi-conduc-tor}

\title{An Effective Weight Initialization Method for Deep Learning: Application to Satellite Image Classification}

  \author{Wadii~Boulila,
Eman~Alshanqiti,
         Ayyub~Alzahem,

         Anis~Koubaa,
         and~Nabil Mlaiki
\thanks{W. Boulila, A. Alzahem, and A. Koubaa are with the Robotics and Internet-of-Things Laboratory, Prince Sultan University, Riyadh, Saudi Arabia}
 \thanks{W. Boulila is with the RIADI Laboratory, National School of Computer Sciences, University of Manouba, Manouba, Tunisia}
  \thanks{E. Alshanqiti is with the College of Computer Science and Engineering, Taibah University, Medina, Saudi Arabia}
  \thanks{N. Mlaiki is with the Department of Mathematics and Sciences, Prince Sultan University, Riyadh, Saudi Arabia}}

\maketitle

\begin{abstract}
The growing interest in satellite imagery has triggered the need for efficient mechanisms to extract valuable information from these vast data sources, providing deeper insights. Even though deep learning has shown significant progress in satellite image classification. Nevertheless, in the literature, only a few results can be found on weight initialization techniques. These techniques traditionally involve initializing the networks' weights before training on extensive datasets, distinct from fine-tuning the weights of pre-trained networks. In this study, a novel weight initialization method is proposed in the context of satellite image classification. The proposed weight initialization method is mathematically detailed during the forward and backward passes of the convolutional neural network (CNN) model. Extensive experiments are carried out using six real-world datasets. Comparative analyses with existing weight initialization techniques made on various well-known CNN models reveal that the proposed weight initialization technique outperforms the previous competitive techniques in classification accuracy. The complete code of the proposed technique, along with the obtained results, is available at
https://github.com/WadiiBoulila/Weight-Initialization
\end{abstract}


%
\IEEEpeerreviewmaketitle

\section{Introduction}

\IEEEPARstart{O}{ver} the recent century, remote sensing (RS) has gained growing popularity since RS data plays an invaluable role in many fields such as crop growth tracking, land use or land cover change prediction, disaster monitoring, etc. Satellite images are now used by nations for political decision-making, civil security activities, police, and geographic information systems. All these applications require satellite image classification to extract meaningful information from them.

Satellite image classification refers to arranging pixels into meaningful classes. It can be done using various methods and techniques that can be supervised, unsupervised, or semi-supervised. Abburu and Golla \cite{abburu2015satellite} claimed that neural networks (NN) could replicate the human learning process to connect image pixels with the correct meaningful labels. NN-based algorithms are used in satellite image classification to benefit from the simple integration of additional data into the classification process and enhance classification accuracy.

Selecting appropriate initial weights and activation functions is crucial to prevent the gradient vanishing or exploding problem \cite{narkhede2022review}. Various weight initialization methods have been proposed in different fields to reduce the execution time of deep learning (DL) techniques. Some of these methods include normal initialization, constant initialization, Lecun initialization, random initialization, Xavier initialization, and He initialization. Despite this variety, there are very few published results related to the weight initialization of DL techniques in the context of satellite images.
Nowadays, with the continuous progress in satellite sensors, we have massive satellite image volumes, which the RS community refers to as big data. The challenge is to extract valuable information in the context of RS big data. Classification has emerged as one of the most effective and reliable methods for extracting relevant data from satellite images \cite{dong2021exploring,xue2022self,xu2022unsupervised}. Moreover, RS image classification is used in various applications such as environmental monitoring, land use/cover detection and prediction, tree species in forests, urban planning, etc. \cite{boulila2019top,boulila2022hybrid,alzahem2023improving}. Many DL techniques were developed in the context of satellite image classification \cite{yuan2021review}. Maintaining accuracy while training in a good runtime is problematic for DL approaches. Weight initialization is considered an appropriate step to resolve this issue. It describes how an NN layer's initial weight values are assigned to prevent layer activation outputs from inflating or disappearing.

The primary motivation for conducting this research study is that most existing works on classification focus on developing new DL-based techniques. However, these works disregard the process of weight initialization, which would lead to significant improvements in satellite image classification. Therefore, this research proposes an efficient approach for weight initialization that can help increase the accuracy of DL techniques.
The main contributions of the proposed study are summarized as follows:
\begin{itemize}
\item A novel weight initialization strategy for DL is proposed. A step-by-step mathematical proof and theoretical explanation are provided to detail the newly proposed weight initialization method.
\item Several experiments have been conducted to show the effectiveness of the proposed method on multiple public datasets. Results show excellent performances of the proposed method compared to state-of-the-art related methods. The code of the proposed weight initialization method and the obtained results are shared at https://github.com/WadiiBoulila/Weight-Initialization.
\end{itemize}

Our manuscript is structured as follows. Section 2 discusses related research works. The proposed weight initialization method is discussed in Section 3. Section 4 describes the application of the proposed weight initialization method. Section 5 depicts the experiments conducted on satellite image datasets. The evaluation of the proposed weight initialization method on challenging computer vision dataset is detailed in Section 6. Finally, Section 7 concludes this study and suggests future research perspectives.

\section{Literature review}

In recent years, DL has made significant strides with notable advancements being achieved. Despite the well-known challenges associated with training deep models, some outstanding results have been accomplished. One of the main barriers in training these models comes from identifying the most suitable initialization strategy for the model's parameters. The power of DL relies on its ability to learn features using several hidden layers. Extracted features from the trained model are more abstract and fundamental expressions of the original input data. The input data information can be efficiently reduced by using the unsupervised learning algorithm to accomplish a technique called "layer initialization," which will effectively decrease the depth of the neural network training difficulty. 

In DL, the dataset size and the initial weights play a crucial role. Optimization algorithms (e.g., gradient descent) are used to incrementally change the initial weights to minimize a loss function, which can result in pertinent decisions. Setting initial weights is a starting point for optimization algorithms. Weight initialization aims to speed the convergence time and help establish a stable neural network learning bias. Training the network without a sufficient weight initialization might result in very slow convergence or a failure to converge \cite{deng2020toward}. Furthermore, training the network without proper weight initialization has the potential of creating an inflated or vanishing gradient, which can result in extremely slow convergence or the network failing to converge. When training a network, choosing an appropriate weight initialization approach is crucial \cite{boulila2022weight,ben2022randomly}.

Several weight initialization techniques exist in the literature, such as all-zeros, constant, standard normal, Lecun, random, Xavier, and He \cite{boulila2022weight,mishkin2015all,sussillo2014random,hinton2015distilling,li2020comparison}. Table \ref{tab:methods_comparison} illustrates some essential advantages and limitations of these techniques.

\begin{table*}[htbp]
    \centering
    \caption{Comparison between the most important weight initialization techniques.}
    \renewcommand{\arraystretch}{1.5} 
    \begin{tabular}{p{3.5cm}p{5cm}p{5cm}c}
    \textbf{Initialization Method} & 
    \textbf{Pros} & 
    \textbf{Cons} & 
    \textbf{Ref.}  \\ \hline
    All-zeros initialization & 
    Simplicity & 
    Symmetry problems lead neurons to learn the same features & 
    \cite{kumar2021medical} \\ \hline
    Constant initialization & 
    Simplicity & 
    Symmetry problems lead neurons to learn the same features & 
    \cite{kumar2021medical} \\ \hline
    Standard normal initialization & 
    Even if the back-propagated gradients become lower, the weight gradient variance is approximately constant across layers & 
    When all layers of the same size are assumed, the back-propagated gradient variance will depend on the layer & 
    \cite{pmlr-v9-glorot10a} \\ \hline
    Lecun initialization & 
    Solving growing variance and gradient problems & 
    Ineffective in networks with constant width; the width should grow approximately linearly with the depth to keep this variance bounded & 
    \cite{pmlr-v38-lee15a} \\ \hline
    Random initialization & 
    Increasing accuracy and optimizing the symmetry-breaking procedure. Neurons no longer do the same computation & 
    Leading to a vanishing gradient, a problem with saturation may occur, and the gradient or slope is minimal, resulting in a gradual gradient drop & 
    \cite{kumar2021medical} \\ \hline
    Xavier initialization & 
    Reducing vanishing/exploding chances & 
    Dying neurons during training & 
    \cite{pmlr-v15-glorot11a} \\ \hline
    He initialization & 
    Solving dying neuron problems & 
    Working better for layers with activations of ReLU or LeakyReLU & 
    \cite{He_2015_ICCV} \\ \hline
    ZerO Initialization &
    Solving exploding gradient problem &
    Leading to a vanishing gradient and symmetry problem &
    \cite{zhao2022zero} \\ \hline
    \end{tabular}
    \label{tab:methods_comparison}
\end{table*}

In the recent decade, a growing body of literature has contributed to developing several weight techniques for DL. Based on the best of our knowledge, weight initialization is a very recent topic in RS, and few studies have been published on satellite image classification. In \cite{Kampffmeyer_2016_CVPR_Workshops}, Kampffmeyer et al. proposed three CNN architectures, pixel-to-pixel based and patch-based, for the classification of urban satellite images. The authors analyzed the performance of their approach to small object segmentation. Experiments are conducted using the ISPRS Vaihingen 2D semantic labeling contest dataset. In this paper, the authors have used the He method to initialize the weights or their DL model.

In \cite{8368069}, Kemker et al. suggested a semantic segmentation approach based on a low-shot learning method based on self-taught feature learning. The authors combined self-taught feature learning and semi-supervised classification for multispectral and hyperspectral images. Results are conducted on publicly available hyperspectral images collected by three different NASA sensors and depict a high bar for low-shot learning. In this paper, the authors have initialized their model using Xavier initialization. 

Piramanayagam et al. \cite{rs10091429} described a CNN-based technique for pixel-wise semantic segmentation using information from multisensor RS images. The authors presented an early CNN feature fusion based on various spectral bands. This reduced the amount of computing time and GPU memory needed for training. Four datasets are used in the experiments: IEEE Zeebruges, ISPRS Potsdam, Sentinel-2, Sentinel-1, and Vaihingen. The authors of this research used Xavier initialization to initialize their model.

Wang et al. demonstrated in \cite{rs12020207} that the U-Net model could partition crops using tiny numbers of weakly supervised labels (i.e., labels of single geotagged points and image-level labels). CNNs may provide accurate segmentation with little supervision, outperforming pixel-level 
techniques such as support vector machines, random forest, and logistic regression. Experiments are carried out utilizing Landsat satellite images from the US Geological Survey. The authors of this research used Xavier initialization to initialize their model.

Zhao et al. \cite{doi:10.1080/01431161.2021.1938738} developed a fuzzy CNN-based model, called RSFCNN, for the semantic segmentation of satellite images. The proposed model learns comprehensive information at the pixel level by extracting features and then conducting fuzzy processing. The fuzzy logic is used to assist CNN in better describing the uncertainty of RS data. Experiments are carried out on two datasets from the semantic labeling contest of ISPRS and CCF Satellite Imagery for AI Classification and Recognition Challenge.

Xia et al. introduced a CNN-based model dubbed DDLNet in \cite{rs13163083}, which is based on edge guidance, deep multiscale supervision, and full-scale skip connection. The authors aim to tackle the edge discontinuity and polygon shape created by classification problems. Experiments are conducted using two high-resolution RS images, one from Google images and one aerial image representing building areas. The authors of this study initialized the DDLNet weights using the weights of a ResNet34 model using ImageNet.

In \cite{su2022improved}, Su et al. suggested improving U-Net using an end-to-end deep CNN combining the DeconvNet, U-Net, DenseNet, and dilated convolution. The idea of using the fusion of the previous techniques is to reduce model parameters, speed up the segmentation runtime, and enhance the segmentation quality. Experiments are conducted using the Potsdam orthophoto dataset. In this paper, the authors have initialized the weights of their model using the He initialization method.

Pan et al. in \cite{pmlr-v162-pan22b} presented a novel approach to weight initialization for Tensorial Convolutional Neural Networks (TCNNs). This was developed in response to the ineffectiveness of traditional Xavier and He initialization methods when applied to TCNNs. Their method successfully generated appropriate weights for the TCNNs and enhanced the accuracy of popular datasets such as CIFAR-10 and Tiny-ImageNet. 

In \cite{zhao2022zero}, Zhao et al. proposed ZerO initialization consisting of only zeros and once. It has been tested using the ResNet-18 model on the CIFAR-10 dataset and ResNet-50 on the ImageNet dataset. ZerO initialization successfully reduced the test error rate by 0.03 to 0.08 std.

In \cite{gadiraju2023application}, Gadiraju et al. discussed the challenges of using transfer learning for crop classification with aerial imagery. Results showed that using the network weights as initial weights for training on the RS dataset or freezing the early layers of the network improves performance compared to training the network from scratch, which was done using random initialization.

In \cite{noman2023remote}, Noman et al. introduced a new approach for change detection using transformers, which achieves state-of-the-art performance on four benchmarks. The method used shuffled sparse-attention and change-enhanced feature fusion to enhance relevant semantic changes and suppress noisy ones. 

By investigating the literature, we can note that initializing the appropriate weights is very important for the training, especially when dealing with complex datasets. Selecting the appropriate weight initializers will improve the performance of the DL models \cite{fong2018meta}. Determining the best way to initialize weights remains a challenge in research. While many studies use established methods like random, Xavier, and He initialization, fewer focus on developing new strategies for choosing the most effective weights.

\section{Proposed Weight Initialization Approach}

\subsection{Description of the Main Steps of the Proposed Approach}

The proposed weight initialization technique improves the training of CNN models specifically for satellite image classification tasks. The main objective of the proposed technique is to initialize the weights of the CNN layers to ensure better classification performance and more efficient learning.

\begin{figure*}[h]
    \centering
    \includegraphics[width=\textwidth]{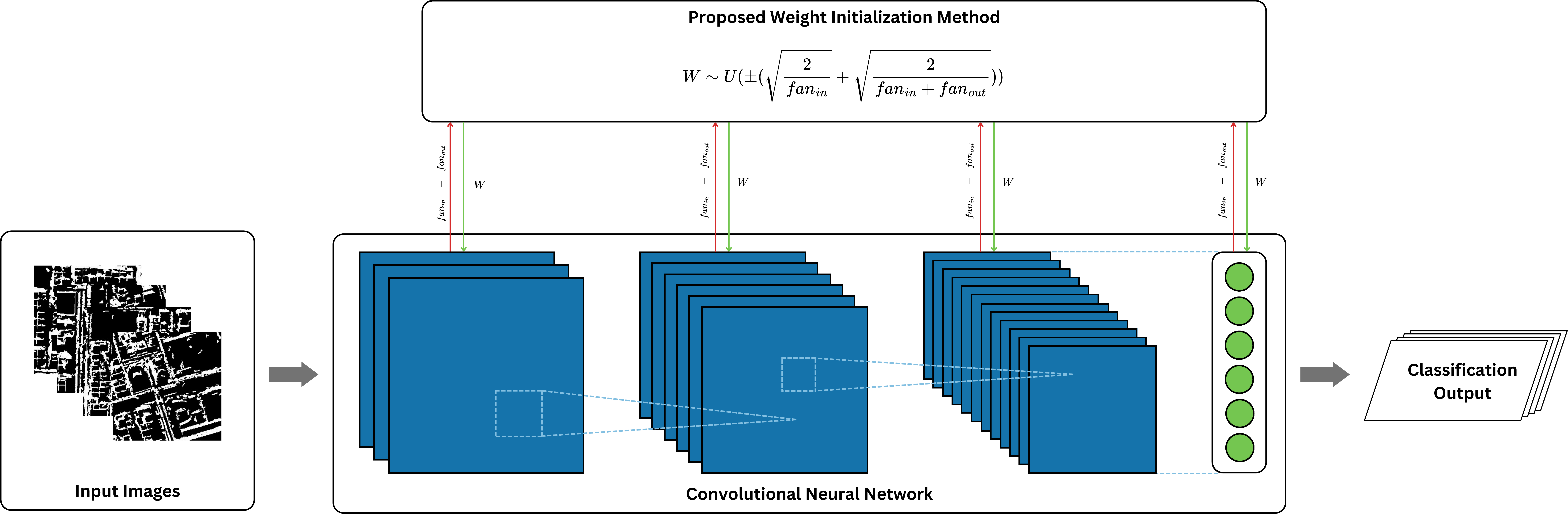}
    \caption{Main steps of the proposed approach.}

    \label{fig:weight_init_cnn}
\end{figure*}

In the proposed approach presented in Figure \ref{fig:weight_init_cnn}, there is a bidirectional interaction with the proposed weight initialization block for each layer. A red arrow from each layer to the weight initialization block carries the $fan_{in}$ and $fan_{out}$ parameters, depicting the need for initializing weights based on these values. These parameters describe the number of input and output connections to a layer, respectively. Subsequently, a green arrow from the weight initialization block back to each layer conveys the initialized weights, denoted by $W$. This cyclical process ensures that each layer's weights are optimally set to facilitate effective learning.

Following the initialization of weights, the CNN undergoes training on the satellite images. This training phase leverages the pre-initialized weights to adjust and fine-tune the network based on the input data and the learning objective, which in this context is classifying satellite images into predetermined categories.

Although the architecture presented here is simplified to illustrate the application of the weight initialization method, it is important to note that this method is applicable to more complex architectures such as ResNet152, VGG19, and MobileNetV2.

Algorithm 1 depicts the main steps of applying the proposed weight initialization to a DL model. Lines 2 and 3 specify the target model and load it. Line 4 loops over the model's modules or layers. Line 5 ensures the next operation applies only to the Linear and Convolution layers. Line 6 is for getting the number of input neurons ($fan_{in}$) and the number of output neurons ($fan_{out}$) in the current layer in the loop. Lines 7 and 8 calculate the proposed weight initialization method from the $fan_{in}$ and $fan_{out}$ of the current layer and compute the uniform distribution from the result value.

\begin{algorithm}
\label{algo:apply_w}
\caption{Initializing proposed weights for a DL model}
\begin{algorithmic}[1]
    \State \textbf{Begin}
    \State $modelName \leftarrow resnet152$ 
    \State $model \leftarrow LoadModel(modelName)$ 
    \For{\textbf{each} $module \in model.modules$}
        \If{$module = Linear$ OR $module = Conv2d$}
            \State $fan_{in}, fan_{out} \leftarrow GetFans(module.weight)$ 
            \State $value \leftarrow \sqrt{\frac{2}{fan_{in}+fan_{out}}} + \sqrt{\frac{2}{fan_{in}}}$
            \State Uniform(-value, value)
        \EndIf
    \EndFor
    \State \textbf{End}
\end{algorithmic} 
\end{algorithm}

\subsection{Mathematical Formulation of the Proposed Weight Initialization Method}

In this section, we present a detailed proof and description of the forward and backward passes of the proposed method. Also, we present the steps for applying the proposed method for the CNN models. The uniform distribution has been selected to keep variance similar across all layers of the CNN model. In this study, we will consider the following assumptions:
\begin{itemize}
    \item \textit{Assumption 1}: We consider that all inputs, weights, and layers are independent and identically distributed.
    \item \textit{Assumption 2}: The weights are initialized with a mean of zero to ensure that the activations have zero means and prevent vanishing or exploding gradients.
    \item \textit{Assumption 3}: The variance of the weights is adjusted based on the number of inputs to each neuron, which helps to keep the signal magnitude consistent across layers.

\end{itemize}

\begin{figure}
    \centering
    \includegraphics[width=3.5in]{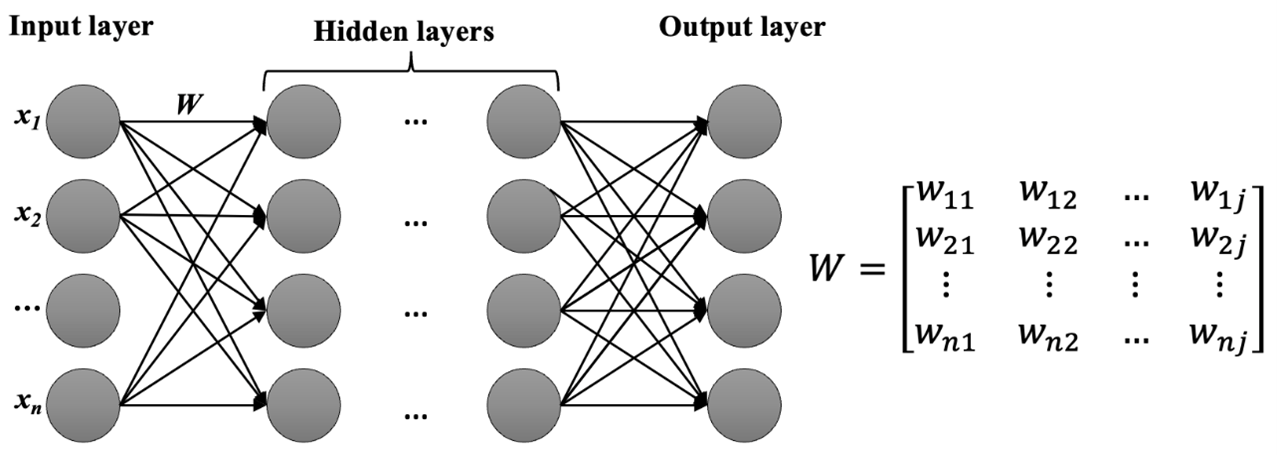}
    \caption{Illustration of the weight initialization process for deep learning networks, where $W$ represents the weights being initialized.}

    \label{fig:weight_init_process}
\end{figure}

Although some assumptions may not fully apply to input data due to intrinsic data characteristics, our initialization strategy is designed to closely adhere to these assumptions. This approach establishes a well-balanced and efficacious foundation for the starting point of model training.
Figure \ref{fig:weight_init_process} depicts the weight initialization process, where $W$ denotes the weights of the DL network.

\subsubsection{Forward Pass}

\begin{figure}
    \centering
    \includegraphics[width=1.5in]{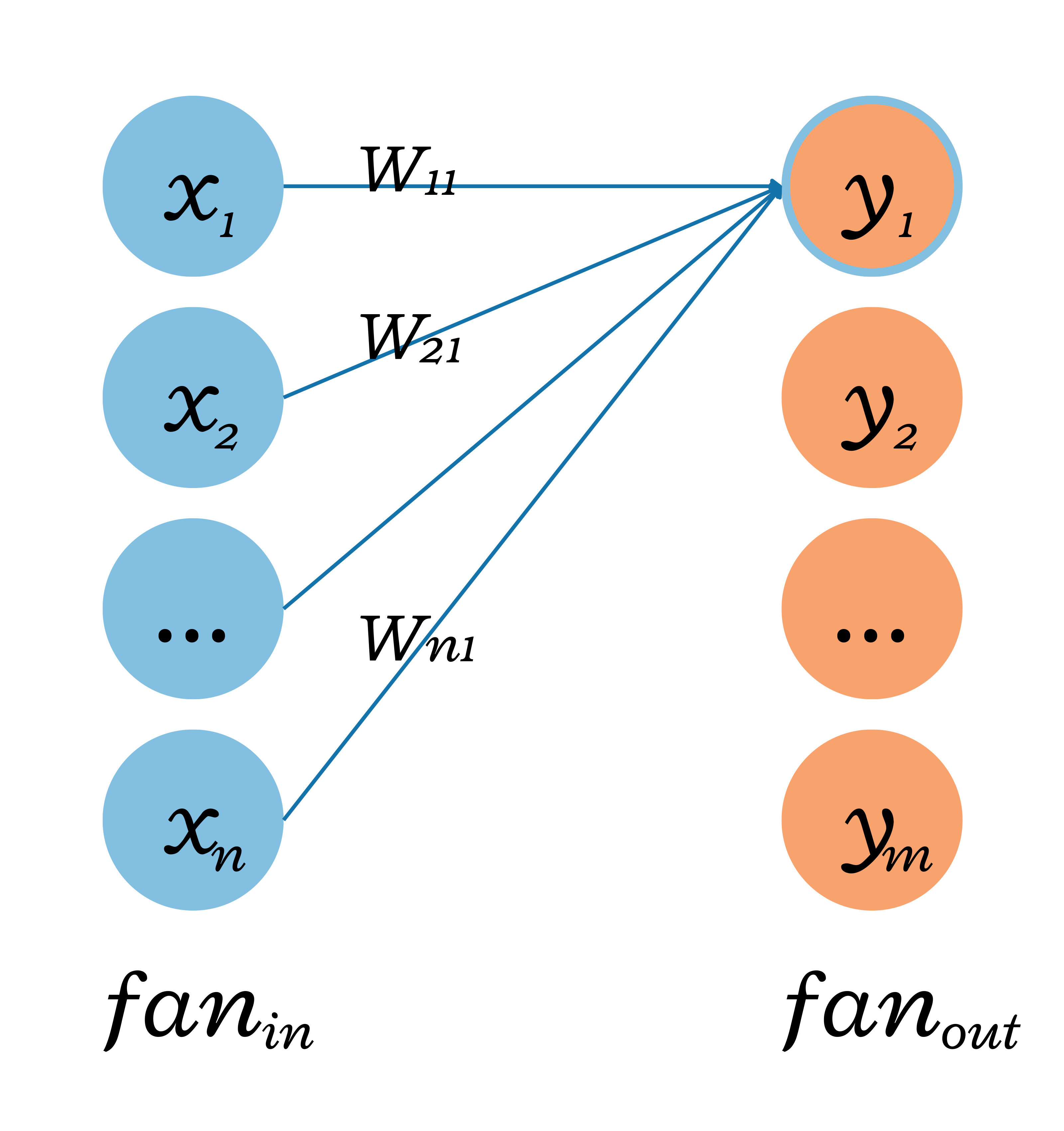}
    \caption{Illustration of the forward pass process, focusing on the activation of unit $y_{1}$ within the network.}

    \label{fig:forward_pass}
\end{figure}

To better explain the forward pass case, we will be singling out one unit $y_{1}$ as depicted in Figure \ref{fig:forward_pass}.

Let us consider that the first hidden layer $(fan_{in})$ weights are $$W=(w_{11}, w_{21}, w_{31},w_{41},\ldots, w_{n1})^{t},$$ $$X=(x_{1}, x_{2}, \ldots, x_{n})^{t}$$ are the input parameters, and $$Y=(y_{1}, y_{2}, \ldots, y_{m})^{t}$$ are the output of the $fan_{in}$.\newline
Assuming that $X$ and $W$ are independent and identically distributed, $y_{1}$ is presented by Equation \ref{eq:w1}.

\begin{equation}
  y_{1} = \sum_{i=1}^{n}x_{i}w_{i1}+b_{1}
  \label{eq:w1}
\end{equation}

We can calculate the variance of $y_{1}$ using the following Equations \ref{eq:w2}, \ref{eq:w4}, and \ref{eq:w5}:

Using Assumption 2 and Assumption 3, we have $ X\bot W,$ and $b$ is constant which leads us to deduce that
\begin{equation}
  Var[y_1] = Var[\sum_{i=1}^{n}x_{i}w_{i1}]=\sum_{i=1}^{n}Var[x_{i}w_{i1}]
  \label{eq:w2}
\end{equation}
Considering \textit{Assumption 1}, by utilizing the independence of $w$ and $x$, we exploit their separate nature to convert the variance of the sum into a summation of individual variances.
Thus, using the fact that $\mathds{E}[x_{i}]= \mathds{E}[w_{i1}]=0$ we deduce that
\begin{align*}\label{w3}
  Var[y_1] &=\sum_{i=1}^{n}Var[x_{i}w_{i1}]\\
  &=\sum_{i=1}^{n} \mathds{E}[x_{i}]^{2}Var[w_{i1}]\\&+\mathds{E}[w_{i1}]^{2}Var[x_{i}]+Var[x_{i}]Var[w_{i1}]\\
  &=\sum_{i=1}^{n}Var[x_{i}]Var[w_{i1}].
\end{align*}
Hence,
\begin{equation}
Var[y_1] = \sum_{i=1}^{n} Var[x_i]Var[w_{i1}]
  \label{eq:w4}
\end{equation}
Now, since by Assumption 1 we have all layers are independent, we can easily deduce that
\begin{equation}
  Var(y_1) = n * Var[x_i]Var[w_{i1}]
  \label{eq:w5}
\end{equation}

The fundamental objective is to maintain variance consistent across all levels. As a result, the variance of $X$ will be equal to the variance of $Y$. This may be performed for the single unit $y 1$ by selecting the variance of its linking weights, as shown in Equation \ref{eq:w6}.

\begin{equation}
  Var[y_1] = Var[x_{i}] \Longleftrightarrow  Var[w_{i1}]=\frac{1}{n}
  \label{eq:w6}
\end{equation}

After that, we generalize the previous result to all the connecting weights between hidden layers X and Y. We will obtain the result illustrated by Equation \ref{eq:w7}.

\begin{equation}
n Var[w_{i1}]=1
  \label{eq:w7}
\end{equation}
and that is
\begin{equation}
fan_{in}Var[w_{i1}]=1
\end{equation}

\subsubsection{Backward Pass (Backpropagation)}

\begin{figure}
    \centering
    \includegraphics[width=1.5in]{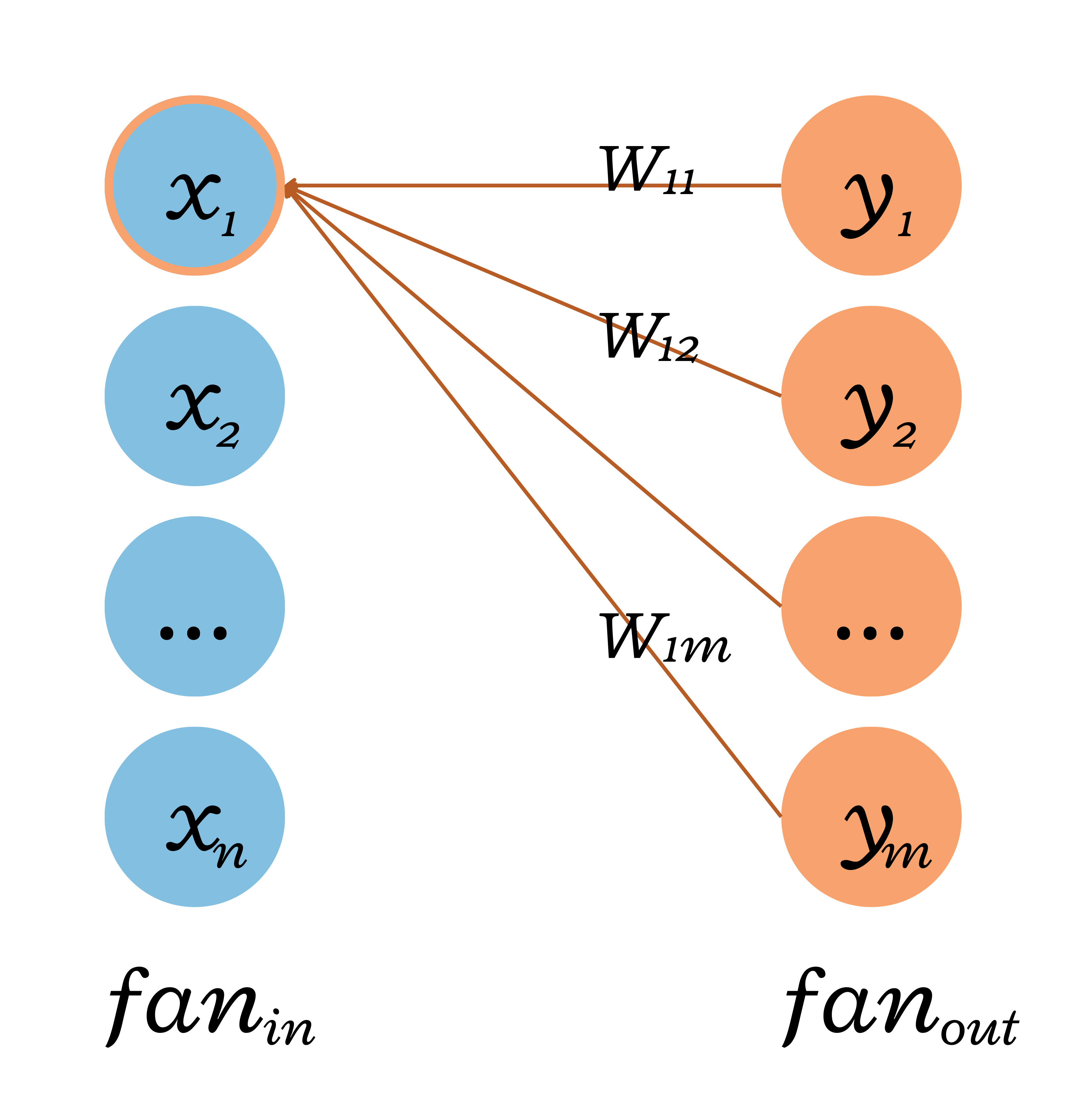}
    \caption{Illustration of the backward pass process, with a focus on the unit $x_{1}$ to elucidate the proposed weight initialization impact.}

    \label{fig:backward_pass}
\end{figure}

For the backward pass, we will also consider the case of one-unit x1 to better explain the proposed weight initialization process, as depicted in Figure \ref{fig:backward_pass}.

We will calculate the variance of the gradients of the unit x1. Mainly, we will make the same assumptions and follow the same steps as illustrated in the forward pass. The gradient of $x_1$ is calculated using equation 10, and its variance is calculated using Equations \ref{eq:w11} and \ref{eq:w12}.

\begin{equation}
  \Delta x_1 = \sum_{i=1}^{m}\Delta y_{i}w_{i1} 
  \label{eq:w10}
\end{equation}

\begin{equation}
  Var[\Delta x_1] = Var[\sum_{i=1}^{m}\Delta y_{i}w_{i1}]
  \label{eq:w11}
\end{equation}

\begin{equation}
  Var[\Delta x_1] =\sum_{i=1}^{m}Var[\Delta y_{i}w_{i1}]
  \label{eq:w12}
\end{equation}
Note that, $\mathds{E}[\Delta y_{i}]= \mathds{E}[w_{i1}]=0$
\begin{align*}\label{w13}
  Var[y_1] &=\sum_{i=1}^{m}Var[\Delta y_{i}w_{i1}]\\
  &=\sum_{i=1}^{n} \mathds{E}[\Delta y_{i}]^{2}Var[w_{i1}]\\&+\mathds{E}[w_{i1}]^{2}Var[\Delta y_{i}]+Var[\Delta y_{i}]Var[w_{i1}]\\
  &=\sum_{i=1}^{n}Var[\Delta y_{i}]Var[w_{i1}].
\end{align*}

\begin{equation}
  Var[\Delta x_1] = m * Var(\Delta y_j)Var(w_{1j})
  \label{eq:w14}
\end{equation}

To maintain the variance of gradients consistent across all layers, we determine the required variance of its linking weights using Equation \ref{eq:w15}.

\begin{equation}
  Var[\Delta x_1] = Var[\Delta y_{j}] \Longleftrightarrow  Var[w_{i1}]=\frac{1}{m}
  \label{eq:w15}
\end{equation}
and that is
\begin{equation}
fan_{out}Var[w_{i1}]=1
\end{equation}

\subsubsection{Weight Distribution}
By using the results found for the forward and backward pass, we deduce the following for all $i:$
\begin{equation}
fan_{in}Var[W]=1,
\end{equation}
and
\begin{equation}
fan_{out}Var[W]=1,
\end{equation}
Thus, 
\begin{equation}
Var[W](fan_{in}+fan_{out})=2.
\end{equation}
which implies
\begin{equation}
Var[W]=\frac{2}{fan_{in}+fan_{out}}
\end{equation}
1) \textbf{Normal distribution:}\\
$$W\sim N(0,\sigma^{2})\Leftarrow Var[W]=\sigma^{2}.$$
Thus, $$\sigma^{2}=\frac{2}{fan_{in}+fan_{out}}\Leftrightarrow \sigma=\pm\sqrt{\frac{2}{fan_{in}+fan_{out}}}$$
2) \textbf{Uniform distribution:}\\
In our approximation we use the interval $(a,b)$ where\newline
$a=-2\sqrt{\frac{6}{fan_{in}+fan_{out}}}$ and $b= 2\sqrt{\frac{6}{fan_{in}}},$
$$W\sim U(-2\sqrt{\frac{6}{fan_{in}+fan_{out}}},2\sqrt{\frac{6}{fan_{in}}}).$$
Therefore,
\begin{align*}
Var[W]&=\displaystyle{\frac{(2\sqrt{\frac{6}{fan_{in}}}+2\sqrt{\frac{6}{fan_{in}+fan_{out}}})^{2}}{12}}\\
&=\displaystyle{\frac{12(\sqrt{\frac{2}{fan_{in}}}+\sqrt{\frac{2}{fan_{in}+fan_{out}}})^{2}}{12}}\\
&=(\sqrt{\frac{2}{fan_{in}}}+\sqrt{\frac{2}{fan_{in}+fan_{out}}})^{2}
\end{align*}
Thus, $W$ follows the normal distribution with a coefficient
\begin{equation}
W\sim U(\pm(\sqrt{\frac{2}{fan_{in}}}+\sqrt{\frac{2}{fan_{in}+fan_{out}}}))
\end{equation}


In the proposed study, maintaining equal variances between the input and output of each layer is considered. This assumption offers several benefits in deep learning. First, it ensures stable learning dynamics throughout the network, preventing the occurrence of unstable gradients caused by varying variances. Second, a consistent gradient flow is promoted by keeping the input and output variances approximately equal, facilitating effective learning. Additionally, it helps avoid saturation and the issues of vanishing or exploding gradients, which can hinder training. Finally, this balanced variance initialization contributes to efficient training by reducing convergence difficulties and enabling faster and more reliable model learning.


\section{Application of the Proposed Weight Initialization Method}

To evaluate the performance of the proposed weight initialization method, we applied it to well-known DL models.

\section{Experiments on Satellite Image Datasets}

\subsection{Dataset Description}

The dataset utilized in this study consisted of 37,774 satellite images with 2.5 meters of spatial resolution collected by the Spot satellite. Ortho-rectification and spatial registration are used to radiometrically and geometrically rectify the images under consideration. There are four classes in this: road, vegetation, bare soil, and buildings. The quantity of the images per label is shown in table \ref{tab:ds}.

\begin{table}[htbp]
    \centering
    \caption{Dataset labels with the quantity of the images per label}
    \begin{tabular}{@{}|c|c|@{}}
    \hline
    \textbf{Label} & \textbf{Quantity} \\ \hline
    Building & 9730 \\ \hline
    Vegetation & 8440 \\ \hline
    Bare soil & 9124 \\ \hline
    Road & 10480 \\ \hline
    \end{tabular}
    \label{tab:ds}
\end{table}

\begin{figure}
    \centering
    \includegraphics[width=3.5in]{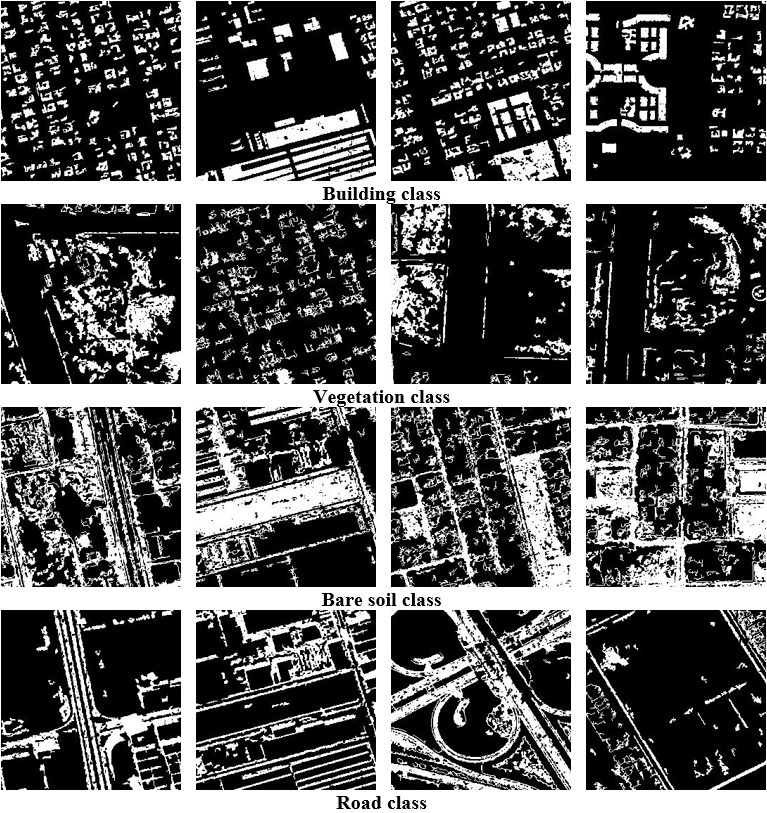}
    \caption{A sample from the satellite image dataset.}

    \label{fig:sample}
\end{figure}

The dataset is randomly split into 60\% (22666 images) for training the model, 20\% (7554 images) for validation, and the remaining 20\% (7554 images) used for testing purposes. The four land cover types are obtained after a semantic segmentation using previous work [2]. Satellite images used in this study have a resolution of 256x256 pixels and are stored in folders labeled with the class name. Figure \ref{fig:sample} shows a sample of this dataset, where white signifies a specific land cover class and black denotes the values of other classes.

\subsection{Results}
In this section, the proposed weight initialization method is applied to three DL models, namely Resnet152V2, VGG19, and MobileNetV2. These DL models have been applied to classify satellite images for the previous dataset. The models were trained for 100 epochs, each consisting of 32 batches. Xavier, He, and the proposed weight initialization method are applied to the three CNN models. All the models are trained at a learning rate of 1e-4 with Adam optimizer.

The classification results of the three models with different weight initialization methods are represented in Table \ref{tab:sat_results}, where column Model represents the used DL model, column Method represents the weight initialization method, P is for precision, R is for recall, F1 is for F1-score, VA is for the validation accuracy, and AA is for the average accuracy. As presented in Table \ref{tab:sat_results}, the proposed method provides better performance than the Xavier and He initialization methods for the three CNN models according to metrics precision, recall, F1-score, validation accuracy, and average accuracy. For the average accuracy, each model has been evaluated ten times, and the average of the achieved validation accuracies has been saved. 

\begin{table}[htbp]
    \centering
    \caption{Satellite Images Classification Report}
    \begin{tabular}{p{1.3cm}p{1cm}p{0.75cm}p{0.75cm}p{0.75cm}p{0.75cm}p{0.75cm}}
    \textbf{Model} & 
    \textbf{Method} & 
    \textbf{P} & 
    \textbf{R} & 
    \textbf{F1} & 
    \textbf{VA} & 
    \textbf{AA} \\ \hline
    \multirow{3}{*}{ResNet152} & 
    He &         0.6161 & 0.6215 & 0.6187 & 0.6299 & 0.6215 \\
    & Xavier &   0.6043 & 0.6120 & 0.6081 & 0.6160 & 0.6120 \\
    & Proposed & \cellcolor[HTML]{AEAEAE}\textbf{0.6152} & \cellcolor[HTML]{AEAEAE}\textbf{0.6232} & \cellcolor[HTML]{AEAEAE}\textbf{0.6191} & \cellcolor[HTML]{AEAEAE}\textbf{0.6345} & \cellcolor[HTML]{AEAEAE}\textbf{0.6232} \\ \hline
    \multirow{3}{*}{VGG19} & 
    He &         0.6432 & 0.6423 & 0.6427 & 0.6560 & 0.6423 \\
    & Xavier &   0.6435 & 0.6463 & 0.6448 & 0.6551 & 0.6463 \\
    & Proposed & \cellcolor[HTML]{AEAEAE}\textbf{0.6581} & \cellcolor[HTML]{AEAEAE}\textbf{0.6574} & \cellcolor[HTML]{AEAEAE}\textbf{0.6577} & \cellcolor[HTML]{AEAEAE}\textbf{0.6574} & \cellcolor[HTML]{AEAEAE}\textbf{0.6574} \\ \hline
    \multirow{3}{*}{MobileNetV2} & 
    He &         0.5994 & 0.6052 & 0.6022 & 0.6299 & 0.6052 \\
    & Xavier &   0.5962 & 0.5975 & 0.5968 & 0.6160 & 0.5975 \\
    & Proposed & \cellcolor[HTML]{AEAEAE}\textbf{0.6038} & \cellcolor[HTML]{AEAEAE}\textbf{0.6081} & \cellcolor[HTML]{AEAEAE}\textbf{0.6059} & \cellcolor[HTML]{AEAEAE}\textbf{0.6345} & \cellcolor[HTML]{AEAEAE}\textbf{0.6081} \\ \hline
    \end{tabular}
    \label{tab:sat_results}
\end{table}

\begin{figure*}[h]
    \centering
    \includegraphics[width=\textwidth]{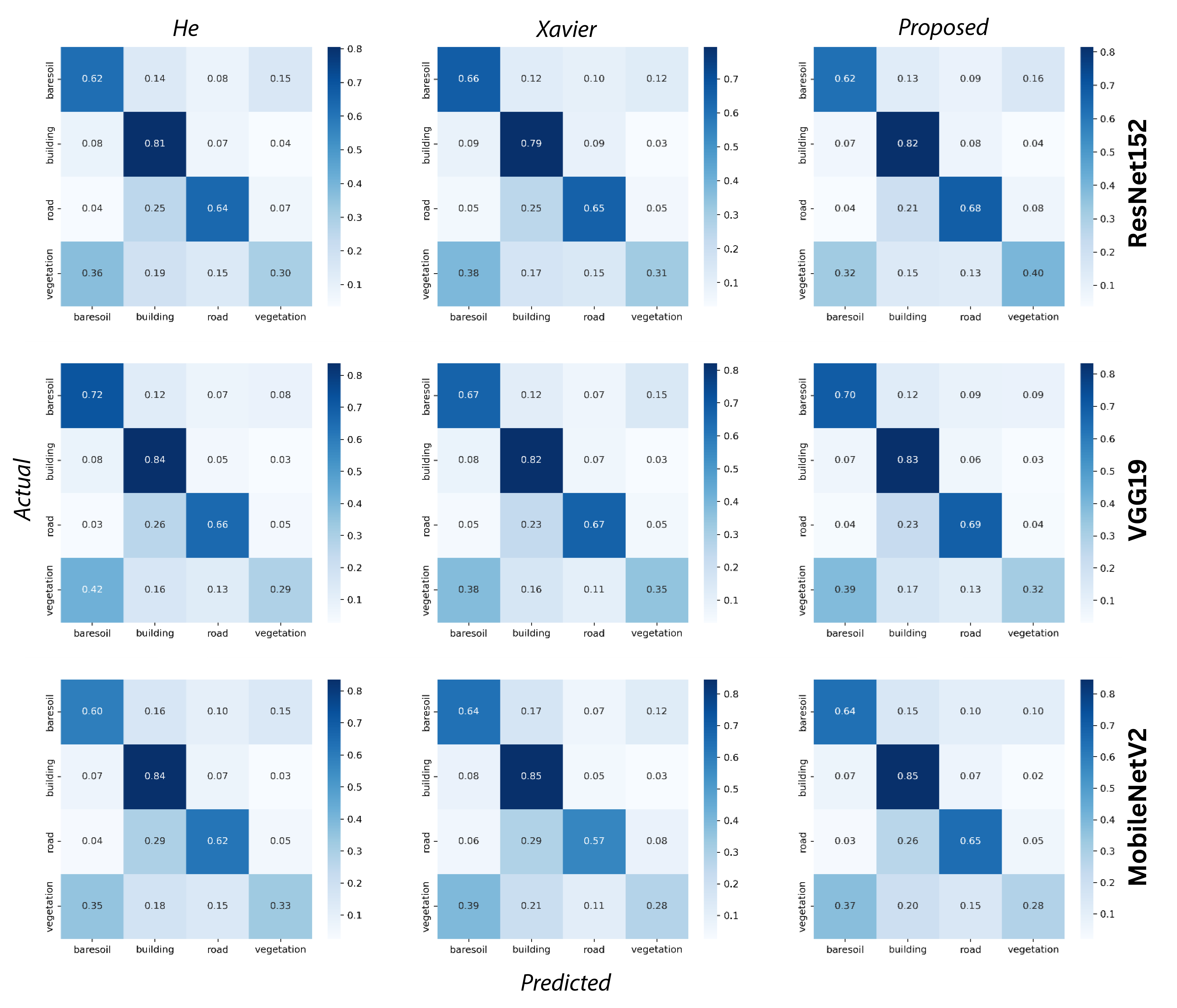}
    \caption{Confusion matrices for ResNet152, VGG19, and MobileNetV2 models, comparing the effectiveness of Xavier, He, and the proposed weight initialization methods in classification accuracy.}
    \label{fig:cm}
\end{figure*}

Figure \ref{fig:cm} depicts the confusion matrix results for ResNet152, VGG19, and MobileNetV2 when applying the three weight initialization techniques, Xavier, He, and the proposed method. Results show that the proposed weight initialization method leads to the highest classification accuracy for all three models compared to Xavier and He methods. Figure \ref{fig:cm}-a) shows classification accuracy for the ResNet152 model. Results show that 59.25\% are correctly classified and 40.75\% are misclassified when applying the He weight initialization, 60.25\% are classified correctly, and 39.75\% are misclassified when applying the Xavier method, and 63\% are classified correctly, and 37\% are misclassified when applying the proposed weight initialization method. Figure \ref{fig:cm}-b) shows classification accuracy for the VGG19 model. Results show that 62.75\% are correctly classified and 37.25\% are misclassified when applying the He weight initialization, 62.75\% are classified correctly, and 37.25\% are misclassified when applying the Xavier method, and 63.5\% are classified correctly, and 36.5\% are misclassified when applying the proposed weight initialization method. Figure \ref{fig:cm}-c) shows classification accuracy for the MobileNetV2 model. Results show that 59.75\% are correctly classified and 40.25\% are misclassified when applying the He weight initialization, 58.5\% are classified correctly, and 41.5\% are misclassified when applying the Xavier method, and 60.5\% are classified correctly, and 39.5\% are misclassified when applying the proposed weight initialization method.

\begin{figure*}
    \centering
    \includegraphics[width=\textwidth]{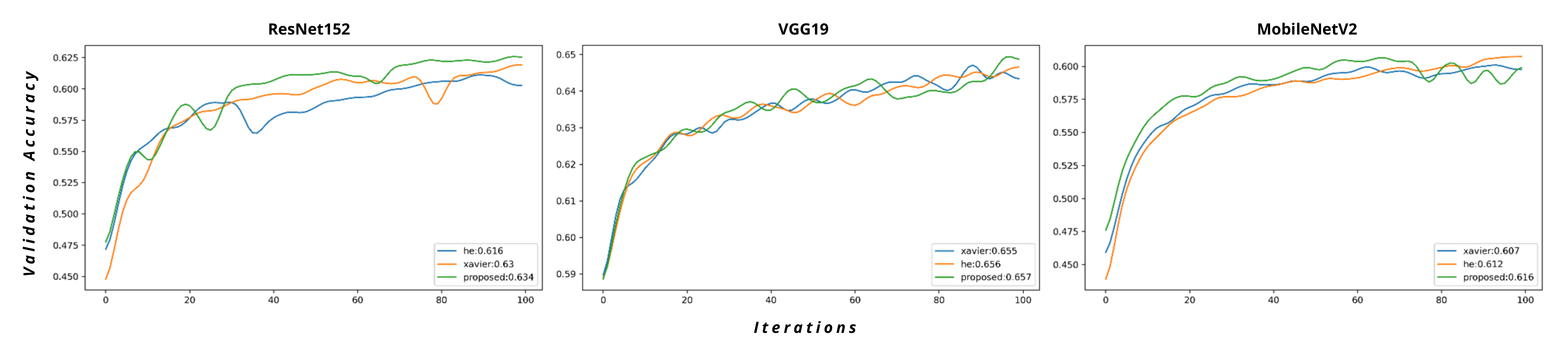}
    \caption{Comparison of validation accuracies over 100 epochs for ResNet152, VGG19, and MobileNetV2 models, demonstrating the performance of the proposed weight initialization method against Xavier and He methods.}

    \label{fig:ca}
\end{figure*}

In addition, the convergence analysis of He, Xavier, and the proposed weight initialization method have been investigated to evaluate the stability of the training pattern and the accuracy they achieve. Figure \ref{fig:ca} depicts the validation accuracy plots for 100 epochs for VGG19, ResNet152, and MobileNetV2. We observe that the validation accuracy of the proposed weight initialization is increasing faster than the validation accuracies in Xavier and He weight initialization methods. The distribution lines in Figure \ref{fig:ca} have been smoothed using the Gaussian filter because they have a very high variation. We note that the proposed weight-initialization method has enhanced the validation accuracy by 0.1\% to 0.4\% compared to He and Xavier methods for the three models, VGG19, ResNet152, and MobileNetV2.

\subsection{Computational Resource Analysis}

As presented in Table \ref{tab:cr}, the metrics under consideration include Allocated Memory, Reserved Memory, and Time, representing the average values computed across all training epochs.

Allocated Memory refers to the amount of GPU memory actively used by the model during the training process. Reserved Memory indicates the total GPU memory reserved by the framework, which is typically higher than the allocated memory to accommodate dynamic memory requirements during training. The Time column reflects the average training duration for completing all epochs in seconds.

A close examination of the table reveals that the computational resources consumed by the proposed weight initialization method are comparable to those of the He and Xavier methods. Specifically, for each model, the differences in allocated and reserved memory among the three methods are minimal, suggesting that the proposed method does not introduce significant computational overhead. Similarly, the training time for each model under different initialization methods is closely aligned, underscoring the efficiency of the proposed method from a computational perspective.

This observation is significant as it implies that the improvements in model accuracy attributed to the proposed weight initialization method do not come at the cost of increased computational resources. Instead, the enhancements in precision, recall, F1-score, validation accuracy, and average accuracy, as outlined in the Results section, are achieved without imposing additional demands on memory allocation or training time.

\begin{table}[htbp]
    \centering
    \caption{Computational Resource Analysis on the Satellite Images Classification}
    \begin{tabular}{p{1.3cm}p{1cm}p{1.3cm}<{\centering}p{1.3cm}<{\centering}p{1.3cm}<{\centering}}
    \textbf{Model} & 
    \textbf{Method} & 
    \textbf{\centering Allocated Memory} & 
    \textbf{\centering Reserved Memory} & 
    \textbf{\centering Time} \\ \hline
    \multirow{3}{*}{ResNet152} & 
    He &         5182  MB & 8313  MB & 362 s \\
    & Xavier &   5181  MB & 8318  MB & 369 s \\
    & Proposed & 5183  MB & 8303  MB & 368 s \\ \hline
    \multirow{3}{*}{VGG19} & 
    He &         4008  MB & 4702  MB & 172 s \\
    & Xavier &   4008  MB & 4633  MB & 171 s \\
    & Proposed & 4008  MB & 4692  MB & 171 s \\ \hline
    \multirow{3}{*}{MobileNetV2} & 
    He &         1926 MB & 2997 MB & 94 s \\
    & Xavier &   1926 MB & 2997 MB & 94 s\\
    & Proposed & 1926 MB & 2976 MB & 94 s \\ \hline
    \end{tabular}
    \label{tab:cr}
\end{table}

\subsection{Evaluation of the Proposed Weight Initialization Method on Public Satellite Datasets}

This section details the performances of the proposed weight initialization method on four public RS datasets, namely, UC-Merced, AID, KSA, and PatternNet.

\subsubsection{RS Public Datasets Description}
The University of California created a dataset called UC-Merced. It is a land use image with 256x256 pixels in size. It contains 2100 RGB images divided equally into 21 classes. The images for numerous metropolitan areas around the country were carefully pulled from massive photographs in the United States Geological Survey National Map \cite{yang2010bag}.

The AID dataset is a large-scale remote sensing images made by assembling common Google Earth photography images. Even though the Google Earth images were post-processed using RGB reconstructions of the original optical aerial photographs. According to research, there is no observable difference between the Google Earth photographs and the genuine optical aerial images, even in mapping land use/cover at the pixel level. Images taken from Google Earth may also be utilized for aerial photography to test scene classification systems. It comprises 10000 photos with a total resolution of 600x600 pixels for all classes\cite{xia2017aid}.

KSA is a multisensor dataset. It was acquired across several cities in the Kingdom of Saudi Arabia (KSA) using three extremely powerful sensors, GeoEye-1, WorldView-2, and IKONOS-2, covering Jeddah, Hufuf, Qassim, Riyadh, and Rajhi farms. This dataset is made up of 13 classes, each comprising 250 photographs with a resolution of 256x256 pixels, \cite{othman2017domain}.

PatternNet dataset is a large remote sensing dataset. It contains 38 classes and 800 256x256 pixel pictures in each class. For several US cities, Google Map API or imagery from Google Earth is used to gather the photos for PatternNet. The classes and associated spatial resolutions are shown in the table below \cite{zhou2018patternnet}.

\subsubsection{Results on RS Public Datasets}
We trained VGG19, ResNet152V2, and MobileNetV2 on UC-Merced, KSA, AID, and PatternNet datasets. All the training was conducted on 100 epochs, 32 batch sizes, and a 0.0001 learning rate. Tables \ref{tab:uc_results},\ref{tab:aid_results}, \ref{tab:ksa_results}, and \ref{tab:pat_results} show the models' evaluation measures for each weight initialization method.

\begin{table}[htbp]
    \centering
    \caption{Performance measures of the DL models on the UC-Merced dataset}
    \begin{tabular}{p{1.3cm}p{1cm}p{0.75cm}p{0.75cm}p{0.75cm}p{0.75cm}p{0.75cm}}
    \textbf{Model} & 
    \textbf{Method} & 
    \textbf{P} & 
    \textbf{R} & 
    \textbf{F1} & 
    \textbf{VA} & 
    \textbf{AA} \\ \hline
    \multirow{3}{*}{ResNet152} & 
    He &         0.4722	& 0.4721	& 0.4721	& 0.5381	& 0.4721 \\
    & Xavier &   0.4431	& 0.4547	& 0.4488	& 0.5095	& 0.4547 \\
    & Proposed & \cellcolor[HTML]{AEAEAE}\textbf{0.4941}	& \cellcolor[HTML]{AEAEAE}\textbf{0.4999}	& \cellcolor[HTML]{AEAEAE}\textbf{0.4969}	& \cellcolor[HTML]{AEAEAE}\textbf{0.5452}	& \cellcolor[HTML]{AEAEAE}\textbf{0.4999} \\ \hline
    \multirow{3}{*}{VGG19} & 
    He &         0.6515	& 0.6457	& 0.6485	& 0.6786	& 0.6457 \\
    & Xavier &   0.6546	& 0.6454	& 0.6499	& 0.6762	& 0.6454 \\
    & Proposed & \cellcolor[HTML]{AEAEAE}\textbf{0.6591}	& \cellcolor[HTML]{AEAEAE}\textbf{0.6523}	& \cellcolor[HTML]{AEAEAE}\textbf{0.6556}	& \cellcolor[HTML]{AEAEAE}\textbf{0.6833}	& \cellcolor[HTML]{AEAEAE}\textbf{0.6523} \\ \hline
    \multirow{3}{*}{MobileNetV2} & 
    He &         0.4058	& 0.4259	& 0.4156	& 0.4500	& 0.4169 \\
    & Xavier &   0.4048	& 0.4169	& 0.4107	& 0.4333	& 0.4259 \\
    & Proposed & \cellcolor[HTML]{AEAEAE}\textbf{0.4190}	& \cellcolor[HTML]{AEAEAE}\textbf{0.4335}	& \cellcolor[HTML]{AEAEAE}\textbf{0.4261}	& \cellcolor[HTML]{AEAEAE}\textbf{0.4690}	& \cellcolor[HTML]{AEAEAE}\textbf{0.4335} \\ \hline
    \end{tabular}
    \label{tab:uc_results}
\end{table}

\begin{table}[htbp]
    \centering
    \caption{Performance measures of the DL models on the AID dataset}
    \begin{tabular}{p{1.3cm}p{1cm}p{0.75cm}p{0.75cm}p{0.75cm}p{0.75cm}p{0.75cm}}
    \textbf{Model} & 
    \textbf{Method} & 
    \textbf{P} & 
    \textbf{R} & 
    \textbf{F1} & 
    \textbf{VA} & 
    \textbf{AA} \\ \hline
    \multirow{3}{*}{ResNet152} & 
    He &         0.3729	& 0.3847	& 0.3787	& 0.3915	& 0.3847 \\
    & Xavier &   0.3916	& 0.4020	& 0.3967	& 0.4140	& 0.4020 \\
    & Proposed & \cellcolor[HTML]{AEAEAE}\textbf{0.3955}	& \cellcolor[HTML]{AEAEAE}\textbf{0.4027}	& \cellcolor[HTML]{AEAEAE}\textbf{0.3990}	& \cellcolor[HTML]{AEAEAE}\textbf{0.4300}	& \cellcolor[HTML]{AEAEAE}\textbf{0.4027} \\ \hline
    \multirow{3}{*}{VGG19} & 
    He &         0.4789	& 0.4824	& 0.4806	& 0.503	    & 0.4824 \\
    & Xavier &   0.4910	& 0.4939	& 0.4924	& 0.507	    & 0.4939 \\
    & Proposed & \cellcolor[HTML]{AEAEAE}\textbf{0.4931}	& \cellcolor[HTML]{AEAEAE}\textbf{0.4972}	& \cellcolor[HTML]{AEAEAE}\textbf{0.4951}	& \cellcolor[HTML]{AEAEAE}\textbf{0.512}	    & \cellcolor[HTML]{AEAEAE}\textbf{0.4972} \\ \hline
    \multirow{3}{*}{MobileNetV2} & 
    He &         0.3079	& 0.3238	& 0.3156	& 0.3510	& 0.3463 \\
    & Xavier &   0.3309	& 0.3463	& 0.3384	& 0.3435	& 0.3238 \\
    & Proposed & \cellcolor[HTML]{AEAEAE}\textbf{0.3402}	& \cellcolor[HTML]{AEAEAE}\textbf{0.3527}	& \cellcolor[HTML]{AEAEAE}\textbf{0.3463}	& \cellcolor[HTML]{AEAEAE}\textbf{0.3575}	& \cellcolor[HTML]{AEAEAE}\textbf{0.3575} \\ \hline
    \end{tabular}
    \label{tab:aid_results}
\end{table}

\begin{table}[htbp]
    \centering
    \caption{Performance measures of the DL models on the KSA dataset}
    \begin{tabular}{p{1.3cm}p{1cm}p{0.75cm}p{0.75cm}p{0.75cm}p{0.75cm}p{0.75cm}}
    \textbf{Model} & 
    \textbf{Method} & 
    \textbf{P} & 
    \textbf{R} & 
    \textbf{F1} & 
    \textbf{VA} & 
    \textbf{AA} \\ \hline
    \multirow{3}{*}{ResNet152} & 
    He &         0.6941	& 0.6947	& 0.6943	& 0.7108	& 0.6947 \\
    & Xavier &   0.7085	& 0.7104	& 0.7094	& 0.7308	& 0.7104 \\
    & Proposed & \cellcolor[HTML]{AEAEAE}\textbf{0.7080}	& \cellcolor[HTML]{AEAEAE}\textbf{0.7138}	& \cellcolor[HTML]{AEAEAE}\textbf{0.7108}	& \cellcolor[HTML]{AEAEAE}\textbf{0.7338}	& \cellcolor[HTML]{AEAEAE}\textbf{0.7138} \\ \hline
    \multirow{3}{*}{VGG19} & 
    He &         0.7990	& 0.7980	& 0.7984	& 0.8292	& 0.7980 \\
    & Xavier &   0.8066	& 0.8044	& 0.8054	& 0.8308	& 0.8044 \\
    & Proposed & \cellcolor[HTML]{AEAEAE}\textbf{0.8161}	& \cellcolor[HTML]{AEAEAE}\textbf{0.8167}	& \cellcolor[HTML]{AEAEAE}\textbf{0.8163}	& \cellcolor[HTML]{AEAEAE}\textbf{0.8400}	& \cellcolor[HTML]{AEAEAE}\textbf{0.8167} \\ \hline
    \multirow{3}{*}{MobileNetV2} & 
    He &         0.6708	& 0.6744	& 0.6725	& 0.6831	& 0.6744 \\
    & Xavier &   0.6672	& 0.6738	& 0.6704	& 0.7031	& 0.6738 \\
    & Proposed & \cellcolor[HTML]{AEAEAE}\textbf{0.6952}	& \cellcolor[HTML]{AEAEAE}\textbf{0.6996}	& \cellcolor[HTML]{AEAEAE}\textbf{0.6973}	& \cellcolor[HTML]{AEAEAE}\textbf{0.7246}	& \cellcolor[HTML]{AEAEAE}\textbf{0.6996} \\ \hline
    \end{tabular}
    \label{tab:ksa_results}
\end{table}

\begin{table}[htbp]
    \centering
    \caption{Performance measures of the DL models on the PatternNet dataset}
    \begin{tabular}{p{1.3cm}p{1cm}p{0.75cm}p{0.75cm}p{0.75cm}p{0.75cm}p{0.75cm}}
    \textbf{Model} & 
    \textbf{Method} & 
    \textbf{P} & 
    \textbf{R} & 
    \textbf{F1} & 
    \textbf{VA} & 
    \textbf{AA} \\ \hline
    \multirow{3}{*}{ResNet152} & 
    He &         0.7410	& 0.7398	& 0.7403	& 0.7298	& 0.7398 \\
    & Xavier &   0.7360	& 0.7266	& 0.7312	& 0.7451	& 0.7266 \\
    & Proposed & \cellcolor[HTML]{AEAEAE}\textbf{0.7790}	& \cellcolor[HTML]{AEAEAE}\textbf{0.7789}	& \cellcolor[HTML]{AEAEAE}\textbf{0.7789}	& \cellcolor[HTML]{AEAEAE}\textbf{0.7896}	& \cellcolor[HTML]{AEAEAE}\textbf{0.7789} \\ \hline
    \multirow{3}{*}{VGG19} & 
    He &         0.8377	& 0.8344	& 0.8360	& 0.8461	& 0.8344 \\
    & Xavier &   0.8324	& 0.8300	& 0.8311	& 0.8362	& 0.8300 \\
    & Proposed & \cellcolor[HTML]{AEAEAE}\textbf{0.8387}	& \cellcolor[HTML]{AEAEAE}\textbf{0.8380}	& \cellcolor[HTML]{AEAEAE}\textbf{0.8383}	& \cellcolor[HTML]{AEAEAE}\textbf{0.8462}	& \cellcolor[HTML]{AEAEAE}\textbf{0.8380} \\ \hline
    \multirow{3}{*}{MobileNetV2} & 
    He &         0.7356	& 0.7273	& 0.7314	& 0.7298	& 0.7273 \\
    & Xavier &   0.7396	& 0.7390	& 0.7392	& 0.7451	& 0.7390 \\
    & Proposed & \cellcolor[HTML]{AEAEAE}\textbf{0.7805}	& \cellcolor[HTML]{AEAEAE}\textbf{0.7802}	& \cellcolor[HTML]{AEAEAE}\textbf{0.7803}	& \cellcolor[HTML]{AEAEAE}\textbf{0.7896}	& \cellcolor[HTML]{AEAEAE}\textbf{0.7802} \\ \hline
    \end{tabular}
    \label{tab:pat_results}
\end{table}

\begin{figure*}
    \centering
    \includegraphics[width=\textwidth]{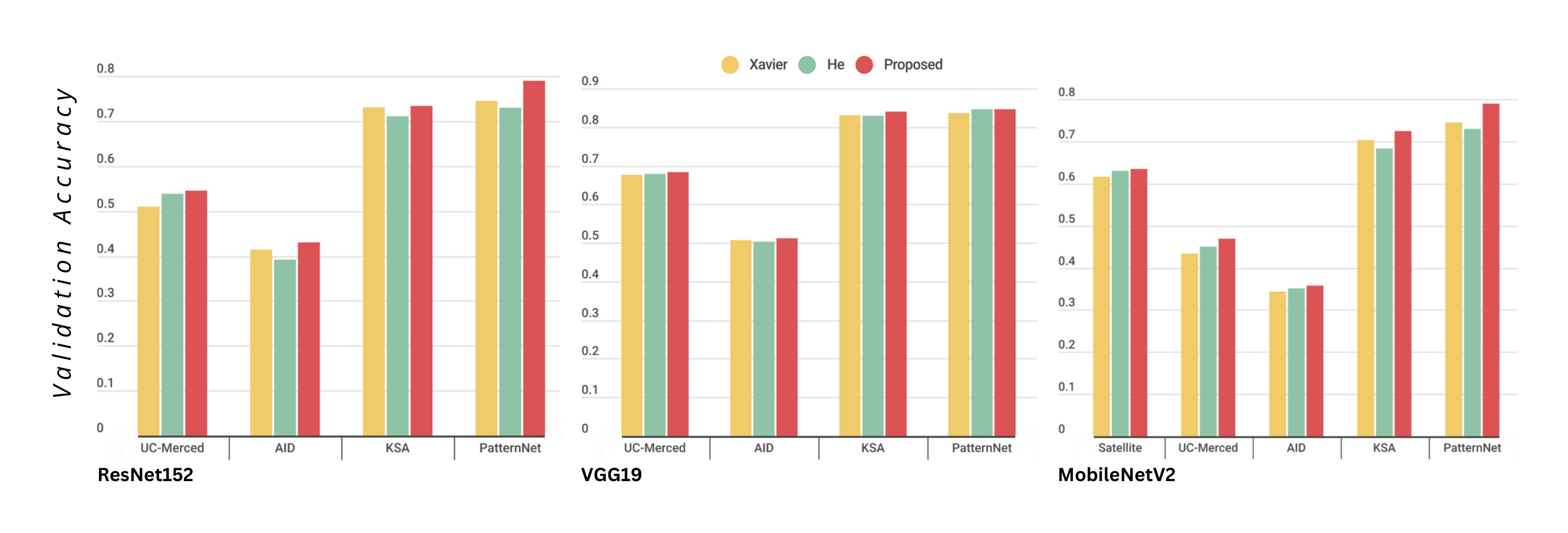}
    \caption{Summary of validation accuracy across different public satellite datasets, highlighting the superior performance of the proposed weight initialization method in bar chart format.}

    \label{fig:satellite_public_bar}
\end{figure*}

We notice that the proposed weight initialization method has achieved the best validation accuracy for all four datasets and for all three models ResNet152, VGG19, and MobileNetV2. Figure \ref{fig:satellite_public_bar} summarizes the validation accuracy of all the presented experiments in a bar chart.

\section{Evaluation of the Proposed Weight Initialization Method on a Non-RS Dataset}

In this section, we extend the evaluation of the proposed weight initialization method to one of the challenging benchmark datasets in the field of computer vision, CIFAR-100. The CIFAR-100 dataset presents a formidable task for image recognition algorithms, consisting of 60,000 color images across 100 fine-grained object classes. Its diverse range of object categories, including animals, vehicles, household items, and natural scenes, demands robust and accurate classification models. To assess the effectiveness of the proposed weight initialization method in such a challenging context, extensive experiments have been conducted on the CIFAR-100 dataset. The results obtained from these experiments are presented in Table \ref{tab:cifar100_results}, providing insights into the performance and a comparative analysis of the proposed method alongside the widely-used Xavier and He initialization techniques. By examining the impact of these weight initialization methods on the accuracy and convergence of DL models, we aim to advance our understanding of initialization strategies and their application in complex image recognition tasks.

\begin{table}[htbp]
    \centering
    \caption{Performance measures of the DL models on CIFAR-100 dataset}
    \begin{tabular}{p{1.3cm}p{1cm}p{0.75cm}p{0.75cm}p{0.75cm}p{0.75cm}p{0.75cm}}
    \textbf{Model} & 
    \textbf{Method} & 
    \textbf{P} & 
    \textbf{R} & 
    \textbf{F1} & 
    \textbf{VA} & 
    \textbf{AA} \\ \hline
    \multirow{3}{*}{ResNet152} & 
    He &         0.5531	& 0.5508	& 0.5468	& 0.5507	& 0.5508 \\
    & Xavier &   0.4959	& 0.5009	& 0.4926	& 0.4975	& 0.5009 \\
    & Proposed & \cellcolor[HTML]{AEAEAE}\textbf{0.5545}	& \cellcolor[HTML]{AEAEAE}\textbf{0.5542}	& \cellcolor[HTML]{AEAEAE}\textbf{0.5502}	& \cellcolor[HTML]{AEAEAE}\textbf{0.5514}	& \cellcolor[HTML]{AEAEAE}\textbf{0.5542} \\ \hline
    \multirow{3}{*}{VGG19} & 
    He &         0.6724	& 0.6682	& 0.6675	& 0.6690	& 0.6682 \\
    & Xavier &   0.6708	& 0.6658	& 0.6654	& 0.6658	& 0.6658 \\
    & Proposed & \cellcolor[HTML]{AEAEAE}\textbf{0.6765}	& \cellcolor[HTML]{AEAEAE}\textbf{0.6717}	& \cellcolor[HTML]{AEAEAE}\textbf{0.6710}	& \cellcolor[HTML]{AEAEAE}\textbf{0.6737}	& \cellcolor[HTML]{AEAEAE}\textbf{0.6717} \\ \hline
    \multirow{3}{*}{MobileNetV2} & 
    He &         0.5590	& 0.5633	& 0.5560	& 0.5682	& 0.5633 \\
    & Xavier &   0.5563	& 0.5595	& 0.5529	& 0.5652	& 0.5595 \\
    & Proposed & \cellcolor[HTML]{AEAEAE}\textbf{0.5638}	& \cellcolor[HTML]{AEAEAE}\textbf{0.5673}	& \cellcolor[HTML]{AEAEAE}\textbf{0.5608}	& \cellcolor[HTML]{AEAEAE}\textbf{0.5683}	& \cellcolor[HTML]{AEAEAE}\textbf{0.5673} \\ \hline
    \end{tabular}
    \label{tab:cifar100_results}
\end{table}

\begin{figure}
    \centering
    \includegraphics[width=3.5in]{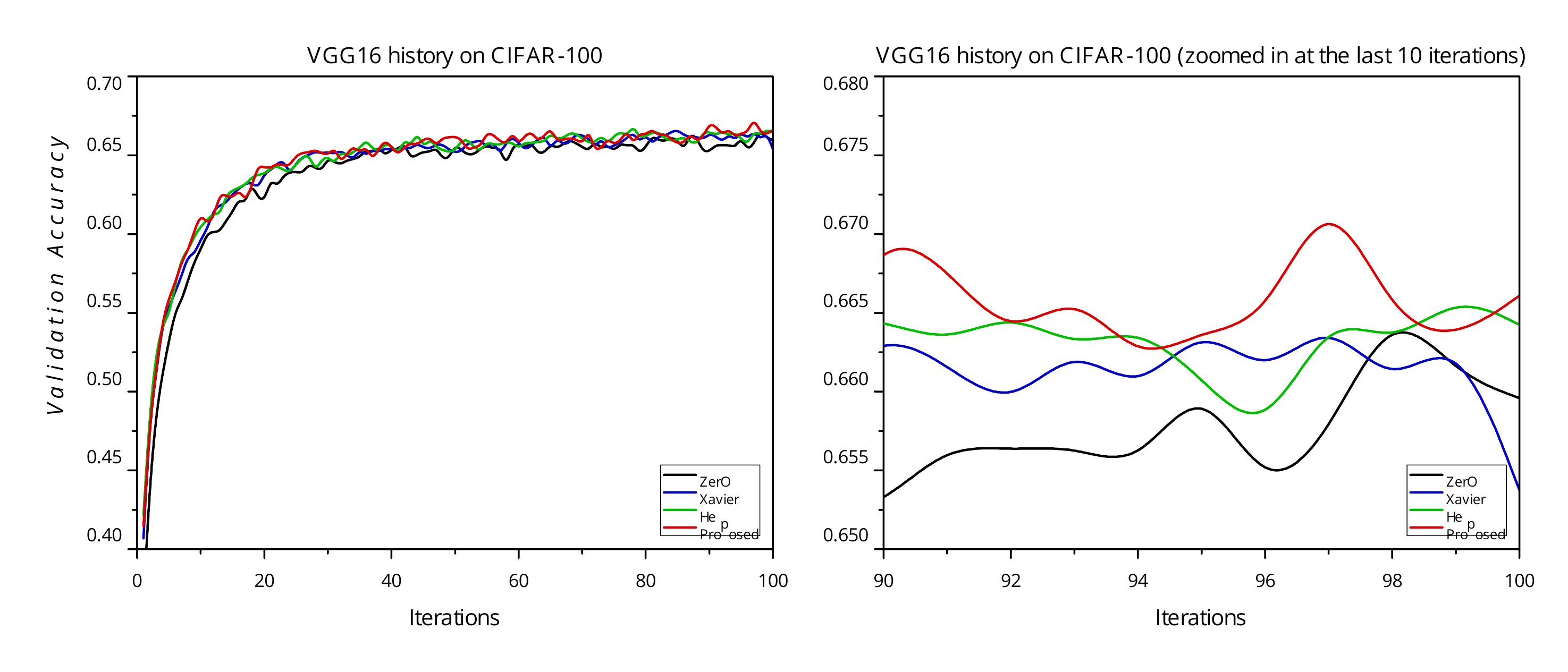}
    \caption{Comparison of validation accuracy during training for our proposed method, Xavier, He, and zerO (\cite{zhao2022zero}) weight initialization methods.} 
    \label{fig:cifar100_accuracy}
\end{figure}

\begin{figure}
    \centering
    \includegraphics[width=3.5in]{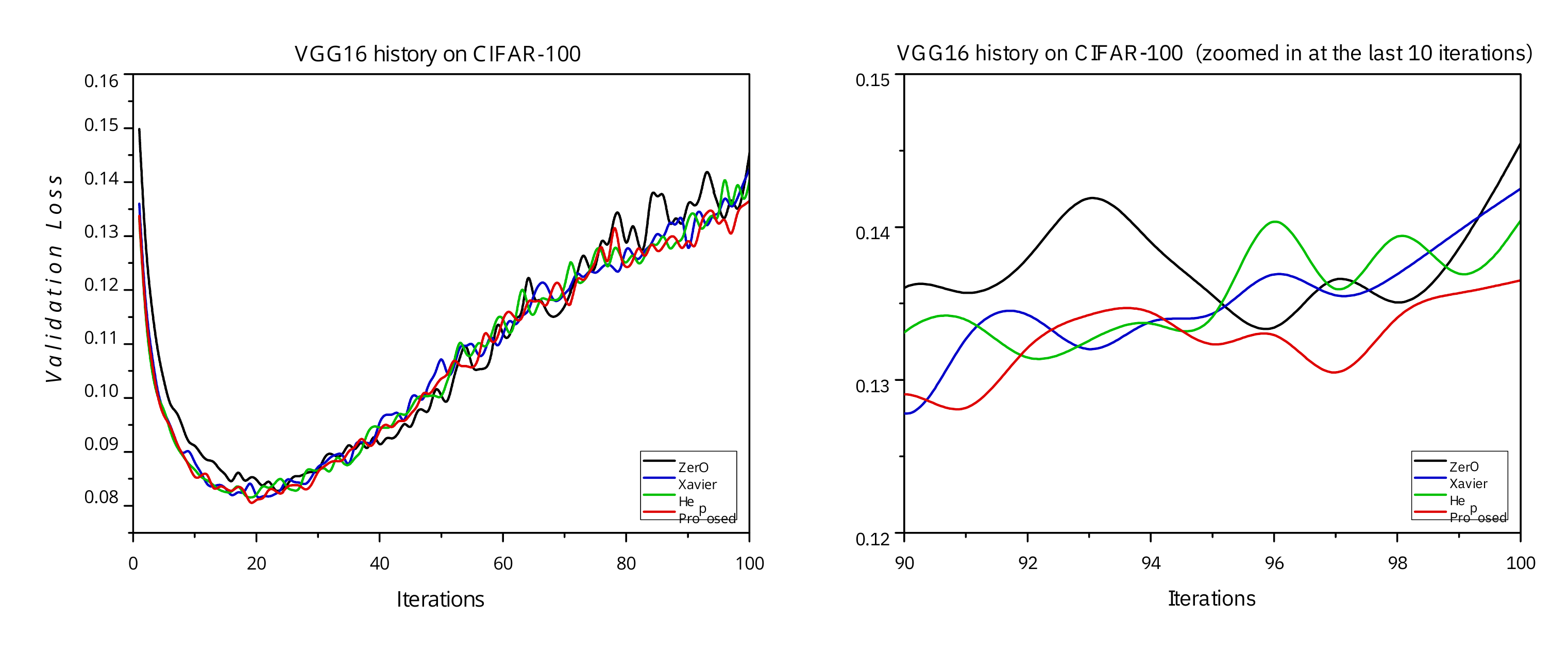}
    \caption{Comparison of validation loss during training for our proposed method, Xavier, He, and zerO (\cite{zhao2022zero}) weight initialization methods.} 
    \label{fig:cifar100_loss}
\end{figure}

The training progress plots in Figure \ref{fig:cifar100_accuracy} and Figure \ref{fig:cifar100_loss} illustrate the performance of the proposed weight initialization method, as well as the Xavier and He, on the CIFAR-100 dataset. Figure \ref{fig:cifar100_accuracy} displays the training progress of validation accuracy, while Figure \ref{fig:cifar100_loss} focuses on validation loss.

The analysis of the plots shows that the proposed weight initialization method outperforms the three other weight initialization techniques in terms of both accuracy and loss, as shown in both the overall training progress and the zoomed-in subplots. The performance advantage of the proposed method is visually apparent, with consistently higher accuracy values and lower loss values throughout the training process.

The comparison with He, Xavier, and zerO initialization methods further confirms the superior performance of the proposed approach. Notably, the zoomed-in subplots highlight the enhanced accuracy and reduced loss achieved by our proposed method in the final ten iterations. These findings highlight the effectiveness of the proposed weight initialization method in improving accuracy and minimizing the discrepancy between predicted and actual values.

\section{Conclusions and Future Works}

This paper details a novel technique of weight initialization for CNN models. The proposed technique is mathematically detailed during the forward and backward passes of the CNN model. Additionally, extensive experiments have been conducted to test and evaluate the performances of the proposed technique with regard to state-of-the-art weight initialization methods. All these techniques were applied to different DL models in the context of satellite image classification. Results show that the proposed weight initialization technique produced the highest precision, recall, and F1-score. Furthermore, the proposed weight initialization method has been evaluated on five public datasets, 4 in the context of RS and 1 in the context of computer vision. Results highlighted good performances of the proposed weight initialization methods. Future research may also investigate the evaluation of the performances of the proposed weight initialization method on the ImageNet dataset.

\ifCLASSOPTIONcaptionsoff
  \newpage
\fi

\bibliographystyle{apalike}
\bibliography{IEEEabrv.bib}

\begin{thebibliography}{}

\bibitem[Abburu and Golla, 2015]{abburu2015satellite}
Abburu, S. and Golla, S.~B. (2015).
\newblock Satellite image classification methods and techniques: A review.
\newblock {\em International journal of computer applications}, 119(8).

\bibitem[Alzahem et~al., 2023]{alzahem2023improving}
Alzahem, A., Boulila, W., Koubaa, A., Khan, Z., and Alturki, I. (2023).
\newblock Improving satellite image classification accuracy using gan-based data augmentation and vision transformers.
\newblock {\em Earth Science Informatics}, 16(4):4169--4186.

\bibitem[Ben~Atitallah et~al., 2022]{ben2022randomly}
Ben~Atitallah, S., Driss, M., Boulila, W., and Ben~Ghezala, H. (2022).
\newblock Randomly initialized convolutional neural network for the recognition of covid-19 using x-ray images.
\newblock {\em International journal of imaging systems and technology}, 32(1):55--73.

\bibitem[Boulila, 2019]{boulila2019top}
Boulila, W. (2019).
\newblock A top-down approach for semantic segmentation of big remote sensing images.
\newblock {\em Earth Science Informatics}, 12(3):295--306.

\bibitem[Boulila et~al., 2022a]{boulila2022weight}
Boulila, W., Driss, M., Alshanqiti, E., Al-Sarem, M., Saeed, F., and Krichen, M. (2022a).
\newblock Weight initialization techniques for deep learning algorithms in remote sensing: Recent trends and future perspectives.
\newblock {\em Advances on Smart and Soft Computing}, pages 477--484.

\bibitem[Boulila et~al., 2022b]{boulila2022hybrid}
Boulila, W., Khlifi, M.~K., Ammar, A., Koubaa, A., Benjdira, B., and Farah, I.~R. (2022b).
\newblock A hybrid privacy-preserving deep learning approach for object classification in very high-resolution satellite images.
\newblock {\em Remote Sensing}, 14(18):4631.

\bibitem[Deng et~al., 2020]{deng2020toward}
Deng, Z., Cao, Y., Zhou, X., Yi, Y., Jiang, Y., and You, I. (2020).
\newblock Toward efficient image recognition in sensor-based iot: a weight initialization optimizing method for cnn based on rgb influence proportion.
\newblock {\em Sensors}, 20(10):2866.

\bibitem[Dong et~al., 2021]{dong2021exploring}
Dong, H., Zhang, L., and Zou, B. (2021).
\newblock Exploring vision transformers for polarimetric sar image classification.
\newblock {\em IEEE Transactions on Geoscience and Remote Sensing}, 60:1--15.

\bibitem[Fong et~al., 2018]{fong2018meta}
Fong, S., Deb, S., and Yang, X.-s. (2018).
\newblock How meta-heuristic algorithms contribute to deep learning in the hype of big data analytics.
\newblock In {\em Progress in intelligent computing techniques: theory, practice, and applications}, pages 3--25. Springer.

\bibitem[Gadiraju and Vatsavai, 2023]{gadiraju2023application}
Gadiraju, K.~K. and Vatsavai, R.~R. (2023).
\newblock Application of transfer learning in remote sensing crop image classification.
\newblock {\em IEEE Journal of Selected Topics in Applied Earth Observations and Remote Sensing}.

\bibitem[Glorot and Bengio, 2010]{pmlr-v9-glorot10a}
Glorot, X. and Bengio, Y. (2010).
\newblock Understanding the difficulty of training deep feedforward neural networks.
\newblock In Teh, Y.~W. and Titterington, M., editors, {\em Proceedings of the Thirteenth International Conference on Artificial Intelligence and Statistics}, volume~9 of {\em Proceedings of Machine Learning Research}, pages 249--256, Chia Laguna Resort, Sardinia, Italy. PMLR.

\bibitem[Glorot et~al., 2011]{pmlr-v15-glorot11a}
Glorot, X., Bordes, A., and Bengio, Y. (2011).
\newblock Deep sparse rectifier neural networks.
\newblock In Gordon, G., Dunson, D., and Dudík, M., editors, {\em Proceedings of the Fourteenth International Conference on Artificial Intelligence and Statistics}, volume~15 of {\em Proceedings of Machine Learning Research}, pages 315--323, Fort Lauderdale, FL, USA. PMLR.

\bibitem[He et~al., 2015]{He_2015_ICCV}
He, K., Zhang, X., Ren, S., and Sun, J. (2015).
\newblock Delving deep into rectifiers: Surpassing human-level performance on imagenet classification.
\newblock In {\em Proceedings of the IEEE International Conference on Computer Vision (ICCV)}.

\bibitem[Hinton et~al., 2015]{hinton2015distilling}
Hinton, G., Vinyals, O., Dean, J., et~al. (2015).
\newblock Distilling the knowledge in a neural network.
\newblock {\em arXiv preprint arXiv:1503.02531}, 2(7).

\bibitem[Kampffmeyer et~al., 2016]{Kampffmeyer_2016_CVPR_Workshops}
Kampffmeyer, M., Salberg, A.-B., and Jenssen, R. (2016).
\newblock Semantic segmentation of small objects and modeling of uncertainty in urban remote sensing images using deep convolutional neural networks.
\newblock In {\em Proceedings of the IEEE Conference on Computer Vision and Pattern Recognition (CVPR) Workshops}.

\bibitem[Kemker et~al., 2018]{8368069}
Kemker, R., Luu, R., and Kanan, C. (2018).
\newblock Low-shot learning for the semantic segmentation of remote sensing imagery.
\newblock {\em IEEE Transactions on Geoscience and Remote Sensing}, 56(10):6214--6223.

\bibitem[Kumar et~al., 2021]{kumar2021medical}
Kumar, A., Dadheech, P., Dogiwal, S., Kumar, S., and Kumari, R. (2021).
\newblock Medical image classification algorithm based on weight initialization-sliding window fusion convolutional neural network.
\newblock In {\em Computer-aided Design and Diagnosis Methods for Biomedical Applications}, pages 193--216. CRC Press.

\bibitem[Lee et~al., 2015]{pmlr-v38-lee15a}
Lee, C.-Y., Xie, S., Gallagher, P., Zhang, Z., and Tu, Z. (2015).
\newblock {Deeply-Supervised Nets}.
\newblock In Lebanon, G. and Vishwanathan, S. V.~N., editors, {\em Proceedings of the Eighteenth International Conference on Artificial Intelligence and Statistics}, volume~38 of {\em Proceedings of Machine Learning Research}, pages 562--570, San Diego, California, USA. PMLR.

\bibitem[Li et~al., 2020]{li2020comparison}
Li, H., Kr{\v{c}}ek, M., and Perin, G. (2020).
\newblock A comparison of weight initializers in deep learning-based side-channel analysis.
\newblock In {\em International Conference on Applied Cryptography and Network Security}, pages 126--143. Springer.

\bibitem[Mishkin and Matas, 2015]{mishkin2015all}
Mishkin, D. and Matas, J. (2015).
\newblock All you need is a good init.
\newblock {\em arXiv preprint arXiv:1511.06422}.

\bibitem[Narkhede et~al., 2022]{narkhede2022review}
Narkhede, M.~V., Bartakke, P.~P., and Sutaone, M.~S. (2022).
\newblock A review on weight initialization strategies for neural networks.
\newblock {\em Artificial intelligence review}, 55(1):291--322.

\bibitem[Noman et~al., 2023]{noman2023remote}
Noman, M., Fiaz, M., Cholakkal, H., Narayan, S., Anwer, R.~M., Khan, S., and Khan, F.~S. (2023).
\newblock Remote sensing change detection with transformers trained from scratch.

\bibitem[Othman et~al., 2017]{othman2017domain}
Othman, E., Bazi, Y., Melgani, F., Alhichri, H., Alajlan, N., and Zuair, M. (2017).
\newblock Domain adaptation network for cross-scene classification.
\newblock {\em IEEE Transactions on Geoscience and Remote Sensing}, 55(8):4441--4456.

\bibitem[Pan et~al., 2022]{pmlr-v162-pan22b}
Pan, Y., Su, Z., Liu, A., Jingquan, W., Li, N., and Xu, Z. (2022).
\newblock A unified weight initialization paradigm for tensorial convolutional neural networks.
\newblock In Chaudhuri, K., Jegelka, S., Song, L., Szepesvari, C., Niu, G., and Sabato, S., editors, {\em Proceedings of the 39th International Conference on Machine Learning}, volume 162 of {\em Proceedings of Machine Learning Research}, pages 17238--17257. PMLR.

\bibitem[Piramanayagam et~al., 2018]{rs10091429}
Piramanayagam, S., Saber, E., Schwartzkopf, W., and Koehler, F.~W. (2018).
\newblock Supervised classification of multisensor remotely sensed images using a deep learning framework.
\newblock {\em Remote Sensing}, 10(9).

\bibitem[Su et~al., 2022]{su2022improved}
Su, Z., Li, W., Ma, Z., and Gao, R. (2022).
\newblock An improved u-net method for the semantic segmentation of remote sensing images.
\newblock {\em Applied Intelligence}, 52(3):3276--3288.

\bibitem[Sussillo and Abbott, 2014]{sussillo2014random}
Sussillo, D. and Abbott, L. (2014).
\newblock Random walk initialization for training very deep feedforward networks.
\newblock {\em arXiv preprint arXiv:1412.6558}.

\bibitem[Wang et~al., 2020]{rs12020207}
Wang, S., Chen, W., Xie, S.~M., Azzari, G., and Lobell, D.~B. (2020).
\newblock Weakly supervised deep learning for segmentation of remote sensing imagery.
\newblock {\em Remote Sensing}, 12(2).

\bibitem[Xia et~al., 2017]{xia2017aid}
Xia, G.-S., Hu, J., Hu, F., Shi, B., Bai, X., Zhong, Y., Zhang, L., and Lu, X. (2017).
\newblock Aid: A benchmark data set for performance evaluation of aerial scene classification.
\newblock {\em IEEE Transactions on Geoscience and Remote Sensing}, 55(7):3965--3981.

\bibitem[Xia et~al., 2021]{rs13163083}
Xia, L., Zhang, J., Zhang, X., Yang, H., and Xu, M. (2021).
\newblock Precise extraction of buildings from high-resolution remote-sensing images based on semantic edges and segmentation.
\newblock {\em Remote Sensing}, 13(16).

\bibitem[Xu et~al., 2022]{xu2022unsupervised}
Xu, H., He, W., Zhang, L., and Zhang, H. (2022).
\newblock Unsupervised spectral--spatial semantic feature learning for hyperspectral image classification.
\newblock {\em IEEE Transactions on Geoscience and Remote Sensing}, 60:1--14.

\bibitem[Xue et~al., 2022]{xue2022self}
Xue, Z., Liu, B., Yu, A., Yu, X., Zhang, P., and Tan, X. (2022).
\newblock Self-supervised feature representation and few-shot land cover classification of multimodal remote sensing images.
\newblock {\em IEEE Transactions on Geoscience and Remote Sensing}, 60:1--18.

\bibitem[Yang and Newsam, 2010]{yang2010bag}
Yang, Y. and Newsam, S. (2010).
\newblock Bag-of-visual-words and spatial extensions for land-use classification.
\newblock In {\em Proceedings of the 18th SIGSPATIAL international conference on advances in geographic information systems}, pages 270--279.

\bibitem[Yuan et~al., 2021]{yuan2021review}
Yuan, X., Shi, J., and Gu, L. (2021).
\newblock A review of deep learning methods for semantic segmentation of remote sensing imagery.
\newblock {\em Expert Systems with Applications}, 169:114417.

\bibitem[Zhao et~al., 2022]{zhao2022zero}
Zhao, J., Schaefer, F.~T., and Anandkumar, A. (2022).
\newblock Zero initialization: Initializing neural networks with only zeros and ones.
\newblock {\em Transactions on Machine Learning Research}.

\bibitem[Zhao et~al., 2021]{doi:10.1080/01431161.2021.1938738}
Zhao, T., Xu, J., Chen, R., and Ma, X. (2021).
\newblock Remote sensing image segmentation based on the fuzzy deep convolutional neural network.
\newblock {\em International Journal of Remote Sensing}, 42(16):6264--6283.

\bibitem[Zhou et~al., 2018]{zhou2018patternnet}
Zhou, W., Newsam, S., Li, C., and Shao, Z. (2018).
\newblock Patternnet: A benchmark dataset for performance evaluation of remote sensing image retrieval.
\newblock {\em ISPRS journal of photogrammetry and remote sensing}, 145:197--209.

\end{thebibliography}


\end{document}